\documentclass{article}

\usepackage{times}
\usepackage{graphicx} 
\usepackage{subfigure} 

\usepackage{amssymb}

\usepackage{natbib}

\usepackage{algorithm}
\usepackage{algorithmic}

\newcommand{\true}{$\surd$}
\newcommand{\false}{$\times$}

\usepackage{hyperref}

\usepackage[accepted]{icml2016}

\icmltitlerunning{Revisiting Semi-Supervised Learning with Graph Embeddings}

\begin{document} 

\twocolumn[
\icmltitle{Revisiting Semi-Supervised Learning with Graph Embeddings}

\icmlauthor{Zhilin Yang}{zhiliny@cs.cmu.edu}
\icmlauthor{William W. Cohen}{wcohen@cs.cmu.edu}
\icmlauthor{Ruslan Salakhutdinov}{rsalakhu@cs.cmu.edu}
\icmladdress{School of Computer Science,
            Carnegie Mellon University}

\icmlkeywords{semi-supervised learning, graph embeddings}

\vskip 0.3in
]

\begin{abstract}

We present a semi-supervised learning framework based on graph embeddings. Given a graph between instances, we train an embedding for each instance to jointly predict the class label and the neighborhood context in the graph. We develop both transductive and inductive variants of our method. In the transductive variant of our method, the class labels are determined by both the learned embeddings and input feature vectors, while in the inductive variant, the embeddings are defined as a parametric function of the feature vectors, so predictions can be made on instances not seen during training.
On a large and diverse set of benchmark tasks, including text classification, distantly supervised entity extraction, and entity classification, we show improved performance over 
many of the existing models. 

\end{abstract}

\section{Introduction}

\textit{Semi-supervised learning} aims to leverage unlabeled data to improve performance. A large number of semi-supervised learning algorithms jointly optimize two training objective functions: the supervised loss over labeled data and the unsupervised loss over both labeled and unlabeled data. \textit{Graph-based semi-supervised learning} defines the loss function as a weighted sum of the supervised loss over labeled instances and a graph Laplacian regularization term \cite{zhu2003semi, zhou2004learning, belkin2006manifold, weston2012deep}. The graph Laplacian regularization is based on the assumption that nearby nodes in a graph are likely to have the same labels. Graph Laplacian regularization is effective because it constrains the labels to be consistent with the graph structure.

Recently developed unsupervised representation learning methods learn embeddings that predict a distributional \textit{context}, e.g.  a word embedding might predict nearby \textit{context words} \cite{mikolov2013distributed, pennington2014glove}, or a node embedding might predict nearby nodes in a graph \cite{perozzi2014deepwalk, tang2015line}.
Embeddings trained with distributional context can be used to boost the performance of related tasks. For example, word embeddings trained from a language model can be applied to part-of-speech tagging, chunking and named entity recognition \cite{collobert2011natural, yang2016multi}.

In this paper we consider not word embeddings but graph embeddings. Existing results show that graph embeddings are effective at classifying the nodes in a graph, such as user behavior prediction in a social network \cite{perozzi2014deepwalk, tang2015line}. However, the graph embeddings are usually learned separately from the supervised task, and hence do not leverage the label information in a specific task. Hence graph embeddings are in some sense complementary to graph Laplacian regularization that does not produce useful features itself and might not be able to fully leverage the distributional information encoded in the graph structure.

The main highlight of our work is to incorporate embedding techniques into the graph-based semi-supervised learning setting. We propose a novel graph-based semi-supervised learning framework, \textit{Planetoid} (Predicting Labels And Neighbors with Embeddings Transductively Or Inductively from Data). The embedding of an instance is jointly trained to predict the class label of the instance and the context in the graph. We then concatenate the embeddings and the hidden layers of the original classifier and feed them to a softmax layer when making the prediction.

Since the embeddings are learned based on the graph structure, the above method is \textit{transductive}, which means we can only predict instances that are already observed in the graph at training time. In many cases, however, it may be desirable to have an \textit{inductive} approach, where predictions can be made on instances unobserved in the graph seen at training time. To address this issue, we further develop an inductive variant of our framework, where we define the embeddings as a parameterized function of input feature vectors; i.e., the embeddings can be viewed as hidden layers of a neural network.

To demonstrate the effectiveness of our proposed approach, we conducted experiments 
on five datasets for three tasks, including text classification, distantly supervised entity extraction, and entity classification. Our inductive method outperforms the second best inductive method by up to $18.7\%$\footnote{$\%$ refers to absolute percentage points thoughout the paper.} points and on average $7.8\%$ points in terms of accuracy. The best of our inductive and transductive methods outperforms the best of all the other compared methods by up to $8.5\%$ and on average $4.1\%$.


\section{Related Work}
\label{sec:related}

\subsection{Semi-Supervised Learning}

\begin{table*}[t]
\caption{Comparison of various semi-supervised learning algorithms and graph embedding algorithms. \true means using the given formulation or information; \false means not available or not using the information. In the column \textit{graph}, \textit{regularization} means imposing regularization with the graph structure; \textit{features} means using graph structure as features; \textit{context} means predicting the graph context.}
\label{tab:comp}
\vskip 0.15in
\begin{center}
\begin{small}
\begin{tabular}{cccccc}
Method & Features & Labels & Paradigm & Embeddings & Graph \\
\hline
\abovespace
TSVM \cite{joachims1999transductive} & \true & \true & Transductive & \false & \false \\
Label propagation \cite{zhu2003semi} & \false & \true & Transductive & \false & Regularization \\
Manifold Reg \cite{belkin2006manifold} & \true & \true & Inductive & \false & Regularization \\
ICA \cite{sen2008collective} & \false & \true & Transductive & \false & Features \\
MAD \cite{talukdar2009new} & \false & \true & Transductive & \false & Regularization \\
Semi Emb \cite{weston2012deep} & \true & \true & Inductive & \true & Regularization \\
Graph Emb \cite{perozzi2014deepwalk} & \false & \false & Transductive & \true & Context \\
Planetoid (this paper) & \true & \true & Both & \true  & Context \\
\end{tabular}
\end{small}
\end{center}
\vskip -0.1in
\end{table*}

Let $L$ and $U$ be the number of labeled and unlabeled instances. Let $\mathbf{x}_{1: L}$ and $\mathbf{x}_{L + 1: L + U}$ denote the feature vectors of labeled and unlabeled instances respectively. The labels $y_{1: L}$ are also given.
Based on both labeled and unlabeled instances, the problem of semi-supervised learning is defined as learning a classifier $f: \mathbf{x} \rightarrow y$. There are two learning paradigms, transductive learning and inductive learning. Transductive learning \cite{zhu2003semi, zhou2004learning} only aims to apply the classifier $f$ on the unlabeled instances observed at training time, and the classifier does not generalize to unobserved instances.
For instance, transductive support vector machine (TSVM) \cite{joachims1999transductive} maximizes the ``unlabeled data margin'' based on the low-density separation assumption that a good decision hyperplane lies on a sparse area of the feature space. Inductive learning \cite{belkin2006manifold, weston2012deep}, on the other hand, aims to learn a parameterized classifier $f$ that is generalizable to unobserved instances.

\subsection{Graph-Based Semi-Supervised Learning}

In addition to labeled and unlabeled instances, a graph, denoted as a $(L + U) \times (L + U)$ matrix $A$, is also given to graph-based semi-supervised learning methods. Each entry $a_{ij}$ indicates the similarity between instance $i$ and $j$, which can be either labeled or unlabeled. The graph $A$ can either be derived from distances between instances \cite{zhu2003semi}, or be explicitly derived from external data, such as a knowledge graph \cite{wijaya2013pidgin} or a citation network between documents \cite{ji2010graph}. In this paper, we mainly focus on the setting that a graph is explicitly given and represents additional information not present in the feature vectors (e.g., the graph edges correspond to hyperlinks between documents, rather than distances between the bag-of-words representation of a document).

Graph-based semi-supervised learning is based on the assumption that nearby nodes tend to have the same labels.
Generally, the loss function of graph-based semi-supervised learning in the binary case can be written as
\begin{eqnarray}
&& \sum_{i = 1}^L l(y_i, f(x_i)) + \lambda \sum_{i, j} a_{ij} \|f(x_i) - f(x_j)\|^2 \nonumber \\
&=& \sum_{i = 1}^L l(y_i, f(x_i)) + \lambda \mathbf{f}^T \Delta \mathbf{f}
\label{eq:graph-obj}
\end{eqnarray}

In Eq. (\ref{eq:graph-obj}), the first term is the standard supervised loss function, where $l(\cdot, \cdot)$ can be log loss, squared loss or hinge loss. The second term is the graph Laplacian regularization, which incurs a large penalty when similar nodes with a large $w_{ij}$ are predicted to have different labels $f(x_i) \not= f(x_j)$. The graph Laplacian matrix $\Delta$ is defined as $\Delta = A - D$, where $D$ is a diagonal matrix with each entry defined as $d_{ii} = \sum_j a_{ij}$. $\lambda$ is a constant weighting factor. (Note that we omit the parameter regularization terms for simplicity.)
Various graph-based semi-supervised learning algorithms define the loss functions as variants of Eq. (\ref{eq:graph-obj}). Label propagation \cite{zhu2003semi} forces $f$ to agree with labeled instances $y_{1: L}$; $f$ is a label lookup table for unlabeled instances in the graph, and can be obtained with a closed-form solution. Learning with local and global consistency \cite{zhou2004learning} defines $l$ as squared loss and $f$ as a label lookup table; it does not force $f$ to agree with labeled instances. Modified Adsorption (MAD) \cite{talukdar2009new} is a variant of label propagation that allows prediction on labeled instances to vary and incorporates node uncertainty. Manifold regularization \cite{belkin2006manifold} parameterizes $f$ in the Reproducing Kernel Hilbert Space (RKHS) with $l$ being squared loss or hinge loss. Since $f$ is a parameterized classifier, manifold regularization is inductive and can naturally handle unobserved instances. 

Semi-supervised embedding \cite{weston2012deep} extends the regularization term in Eq. (\ref{eq:graph-obj}) to be $\sum_{i, j} a_{ij} \|\mathbf{g}(x_i) - \mathbf{g}(x_j)\|^2$,
where $\mathbf{g}$ represents embeddings of instances, which can be the output labels, hidden layers or auxiliary embeddings in a neural network. By extending the regularization from $f$ to $\mathbf{g}$, this method imposes stronger constraints on a neural network.
Iterative classification algorithm (ICA) \cite{sen2008collective} uses a local classifier that takes the labels of neighbor nodes as input,
and employs an iterative process between estimating the local classifier and assigning new labels.

\subsection{Learning Embeddings}

Extensive research was done on learning graph embeddings. A probabilistic generative model was proposed to learn node embeddings that generate the edges in a graph \cite{snijders1997estimation}. A clustering method \cite{handcock2007model} was proposed to learn latent social states in a social network to predict social ties.

More recently, a number of embedding learning methods are based on the Skipgram model, which is a variant of the softmax model. Given an instance and its context, the objective of Skipgram is usually formulated as minimizing the log loss of predicting the context using the embedding of an instance as input features. Formally, let $\{(i, c)\}$ be a set of pairs of instance $i$ and context $c$, the loss function can be written as
\begin{equation}
- \sum_{(i, c)} \log p(c | i) =
- \sum_{(i, c)} \left( \mathbf{w}_c^T \mathbf{e}_i - \log \sum_{c' \in \mathcal{C}} \exp(\mathbf{w}_{c'}^T \mathbf{e}_i) \right)
\label{eq:sg}
\end{equation}
where $\mathcal{C}$ is the set of all possible context, $\mathbf{w}$'s are parameters of the Skipgram model, and $\mathbf{e}_i$ is the embedding of instance $i$. Skipgram was first introduced to learn representations of words, known as word2vec \cite{mikolov2013distributed}. In word2vec, for each training pair $(i, c)$, the instance $i$ is the current word whose embedding is under estimation; the context $c$ is each of the surrounding words of $i$ within a fixed window size in a sentence; the context space $\mathcal{C}$ is the vocabulary of the corpus.
Skipgram was later extended to learn graph embeddings. Deepwalk \cite{perozzi2014deepwalk} uses the embedding of a node to predict the context in the graph, where the context is generated by random walk. More specifically, for each training pair $(i, c)$, the instance $i$ is the current node whose embedding is under estimation; the context $c$ is each of the neighbor nodes within a fixed window size in a generated random walk sequence; the context space $\mathcal{C}$ is all the nodes in the graph. LINE \cite{tang2015line} extends the model to have multiple context spaces $\mathcal{C}$ for modeling both first and second order proximity.

Although Skipgram-like models for graphs have received much recent attention, many other models exist. TransE \cite{bordes2013translating} learns the embeddings of entities in a knowledge graph jointly with their relations. Autoencoders were used to learn graph embeddings for clustering on graphs \cite{tian2014learning}.

\subsection{Comparison}

We compare our approach in this paper with other methods in semi-supervised learning and embedding learning in Table \ref{tab:comp}. 
Unlike our approach, conventional graph Laplacian based methods \cite{zhu2003semi,belkin2006manifold,talukdar2009new} impose regularization on the labels but do not learn embeddings.
Semi-supervised embedding method \cite{weston2012deep} learns embeddings in a neural network, but our approach is different from this method in that instead of imposing regularization, we use the embeddings to predict the context in the graph.
Graph embedding methods \cite{perozzi2014deepwalk,tian2014learning} encode the graph structure into embeddings; however, different from our approach, these methods are purely unsupervised and do not leverage label information for a specific task. Moreover, these methods are transductive and cannot be directly generalized to instances unseen at training time.

\section{Semi-Supervised Learning with Graph Embeddings}
\label{sec:model}

Following the notations in the previous section, the input to our method includes labeled instances $\mathbf{x}_{1: L}$, $y_{1: L}$, unlabeled instances $\mathbf{x}_{L + 1: L + U}$ and a graph denoted as a matrix $A$. Each instance $i$ has an embedding denoted as $\mathbf{e}_i$.

We formulate our framework based on feed-forward neural networks.
Given the input feature vector $\mathbf{x}$, the $k$-th hidden layer of the network is denoted as $\mathbf{h}^k$, which is a nonlinear function of the previous hidden layer $\mathbf{h}^{k - 1}$ defined as:
$
\mathbf{h}^k(\mathbf{x}) = \mbox{ReLU}(\mathbf{W}^k \mathbf{h}^{k - 1}(\mathbf{x}) + b^k),
$
where $\mathbf{W}^k$ and $b^k$ are parameters of the $k$-th layer, and $\mathbf{h}^0(\mathbf{x}) = \mathbf{x}$. We adopt rectified linear unit $\mbox{ReLU}(x) = \max(0, x)$ as the nonlinear function in this work.

The loss function of our framework can be expressed as
\[
\mathcal{L}_s + \lambda \mathcal{L}_u,
\] 
where $\mathcal{L}_s$ is a supervised loss of predicting the labels, and $\mathcal{L}_u$ is an unsupervised loss of predicting the graph context. In the following sections, we first formulate $\mathcal{L}_u$ by introducing how to sample context from the graph, and then formulate $\mathcal{L}_s$ to form our semi-supervised learning framework.


\subsection{Sampling Context}

\begin{figure}[tb]
\vskip 0.2in
\begin{center}
\centerline{\includegraphics[width=0.8\columnwidth]{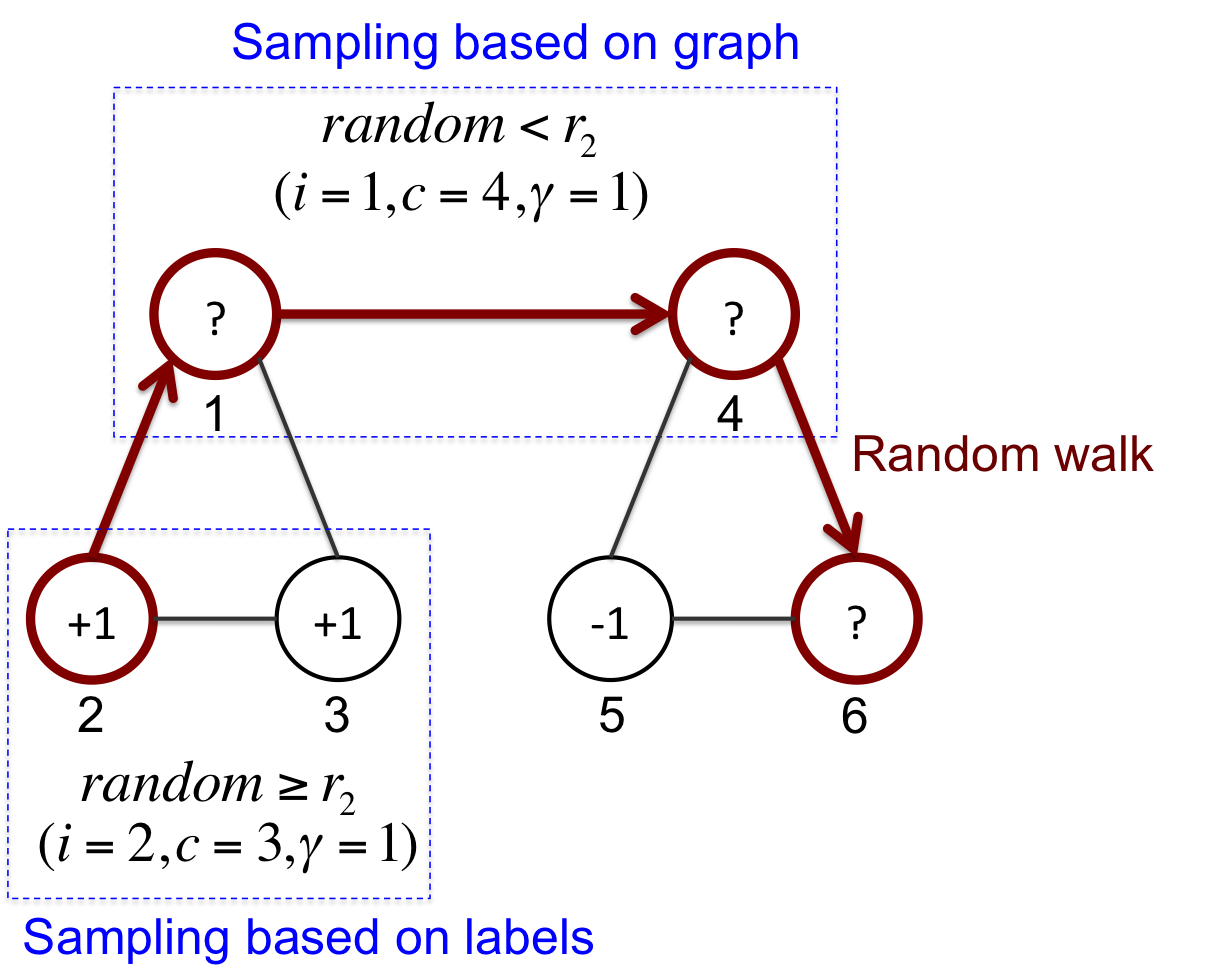}}
\caption{An example of sampling from context distribution $p(i, c, \gamma)$ when $\gamma = 1$ and $d = 2$. In circles, $+1$ denotes positive instances, $-1$ denotes negative instances, and $?$ denotes unlabeled instances. If $random < r_2$, we first sample a random walk $2 \rightarrow 1 \rightarrow 4 \rightarrow 6$, and then sample two nodes in the random walk within distance $d$. If $random \geq r_2$, we sample two instances with the same labels.}
\label{fig:sample}
\end{center}
\vskip -0.2in
\end{figure} 

We formulate the unsupervised loss $\mathcal{L}_u$ as a variant of Eq. (\ref{eq:sg}). Given a graph $A$, the basic idea of our approach is to sample pairs of instance $i$ and context $c$, and then formulate the loss $\mathcal{L}_u$ using the log loss $- \log p(c | i)$ as in Eq. (\ref{eq:sg}). We first present the formulation of $\mathcal{L}_u$ by introducing negative sampling, and then discuss how to sample pairs of instance and context.


It is usually intractable to directly optimize Eq. (\ref{eq:sg}) due to normalization over the whole context space $\mathcal{C}$. Negative sampling was introduced to address this issue \cite{mikolov2013distributed}, which samples negative examples to approximate the normalization term. In our case, we are sampling $(i, c, \gamma)$ from a distribution, where $i$ and $c$ denote instance and context respectively, $\gamma = + 1$ means $(i, c)$ is a positive pair and $\gamma = -1$ means negative. Given $(i, c, \gamma)$, we minimize the cross entropy loss of classifying the pair $(i, c)$ to a binary label $\gamma$:
\[
- \mathbb{I}(\gamma = 1) \log \sigma(\mathbf{w}_c^T \mathbf{e}_i) - \mathbb{I}(\gamma = -1) \log \sigma(- \mathbf{w}_c^T \mathbf{e}_i),
\]
where $\sigma$ is the sigmoid function defined as $\sigma(x) = 1 / (1 + e^{-x})$, and $\mathbb{I}(\cdot)$ is an indicator function that outputs $1$ when the argument is true, otherwise $0$. Therefore, the unsupervised loss with negative sampling can be written as
\begin{equation}
\mathcal{L}_u = - \mathbb{E}_{(i, c, \gamma)} \log \sigma (\gamma \mathbf{w}_c^T \mathbf{e}_i)
\label{eq:sample}
\end{equation}
 The distribution $p(i, c, \gamma)$ is conditioned on labels $y_{1: L}$ and the graph $A$. However, since they are the input to our algorithm and kept fixed, we drop the conditioning in our notation.

We now define the distribution $p(i, c, \gamma)$ directly using a sampling process, which is illustrated in Algorithm \ref{algo:sample}. There are two types of context that are sampled in this algorithm. The first type of context is based on the graph $A$, which encodes the structure (distributional) information, and the second type of context is based on the labels, which we use to inject label information into the embeddings. We use a parameter $r_1 \in (0, 1)$ to control the ratio of positive and negative samples, and use $r_2 \in (0, 1)$ to control the ratio of two types of context.


With probability $r_2$, we sample the context based on the graph $A$. We first uniformly sample a random walk sequence $S$. More specifically, we uniformly sample the first instance $S_1$ from the set $1: L + U$.
Given the previous instance $S_{k - 1} = i$, the next instance $S_k = j$ is sampled with probability $a_{ij} / \sum_{j' = 1}^{L + U} a_{ij'}$.
With probability $r_1$, we sample a positive pair $(i, c)$ from the set $\{(S_j, S_k): |j - k| < d\}$, where $d$ is another parameter determining the window size. With probability $(1 - r_1)$, we uniformly corrupt the context $c$ to sample a negative pair.

With probability $(1 - r_2)$, we sample the context based on the class labels.
Positive pairs have the same labels and negative pairs have different labels. Only labeled instances $1: L$ are sampled.

Our random walk based sampling method is built upon Deepwalk \cite{perozzi2014deepwalk}. In contrast to their method, our method handles real-valued $A$, incorporates negative sampling, and explicitly samples from labels with probability $(1 - r_2)$ to inject supervised information.

An example of sampling when $\gamma = 1$ is shown in Figure \ref{fig:sample}.

\begin{algorithm}[tb]
    \caption{Sampling Context Distribution $p(i, c, \gamma)$}
    \label{algo:sample}
\begin{algorithmic}
    \STATE {\bfseries Input:} graph $A$, labels $y_{1: L}$, parameters $r_1, r_2, q, d$
    \STATE Initialize triplet $(i, c, \gamma)$
    \STATE {\bfseries if} $random < r_1$ {\bfseries then} $\gamma \gets +1$ {\bfseries else} $\gamma \gets -1$
    \IF {$random < r_2$}
        \STATE Uniformly sample a random walk $S$ of length $q$
        \STATE Uniformly sample $(S_j, S_k)$ with $|j - k| < d$
        \STATE $i \gets S_j$, $c \gets S_k$
        \STATE {\bfseries if} $\gamma = -1$ {\bfseries then} uniformly sample $c$ from $1: L + U$
    \ELSE
        \IF {$\gamma = +1$}
            \STATE Uniformly sample $(i, c)$ with $y_i = y_c$
        \ELSE
            \STATE Uniformly sample $(i, c)$ with $y_i \not= y_c$
        \ENDIF
    \ENDIF
    \STATE {\bfseries return} $(i, c, \gamma)$
\end{algorithmic}
\end{algorithm}

\begin{figure}[ht]
\vskip 0.1in
\centering
\subfigure[Transductive Formulation]{\label{fig:trans} \includegraphics[width = 0.8\columnwidth]{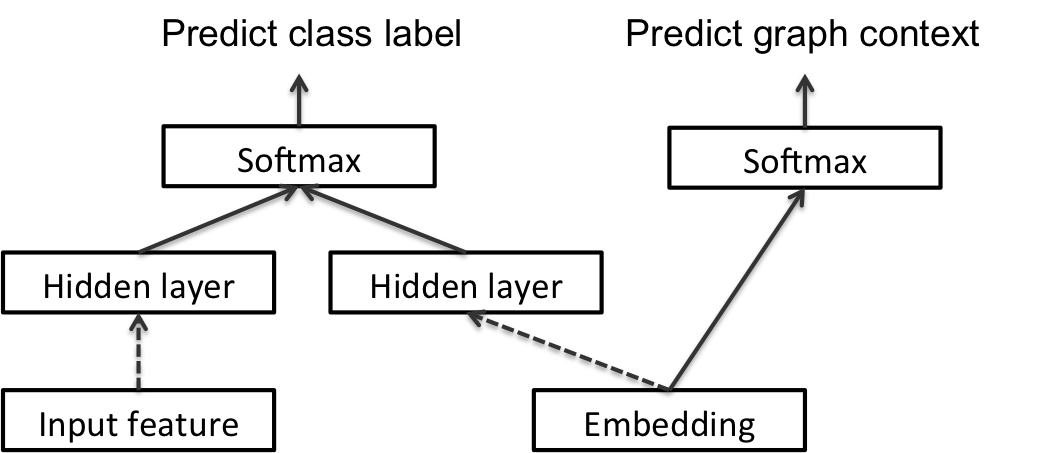}}
\subfigure[Inductive Formulation]{\label{fig:ind} \includegraphics[width = 0.8\columnwidth]{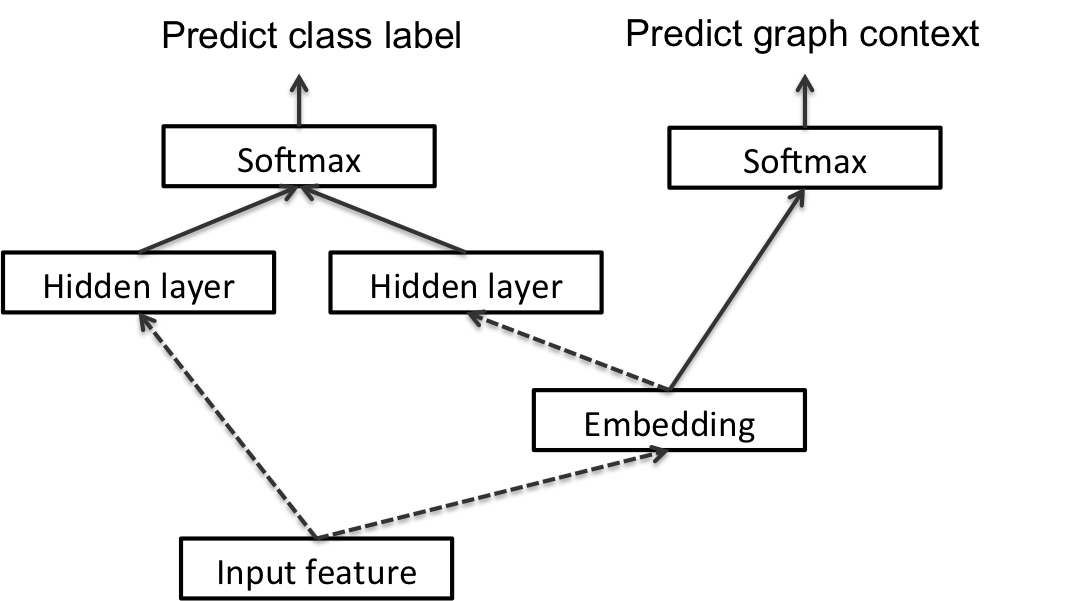}}
\caption{Network architecture: transductive v.s. inductive. Each dotted arrow represents a feed-forward network with an arbitrary number of layers (we use only one layer in our experiments). Solid arrows denote direct connections.}
\label{fig:arch}
\vskip -0.1in
\end{figure} 

\subsection{Transductive Formulation}

In this section, we present a method that infers the labels of unlabeled instances $y_{L + 1: L + U}$ without generalizing to unobserved instances. Transductive learning usually performs better than inductive learning because transductive learning can leverage the unlabeled test data when training the model \cite{joachims1999transductive}.

We apply $k$ layers on the input feature vector $\mathbf{x}$ to obtain $\mathbf{h}^k(\mathbf{x})$, and $l$ layers on the embedding $\mathbf{e}$ to obtain $\mathbf{h}^l(\mathbf{e})$, as illustrated in Figure \ref{fig:trans}. The two hidden layers are concatenated, and fed to a softmax layer to predict the class label of the instance. More specifically, the probability of predicting the label $y$ is written as:
\begin{equation}
p(y | \mathbf{x}, \mathbf{e}) = \frac{\exp [\mathbf{h}^k(\mathbf{x})^T, \mathbf{h}^l(\mathbf{e})^T] \mathbf{w}_y}{\sum_{y'} \exp [\mathbf{h}^k(\mathbf{x})^T, \mathbf{h}^l(\mathbf{e})^T] \mathbf{w}_{y'}},
\label{eq:trans}
\end{equation}
where $[\cdot, \cdot]$ denotes concatenation of two row vectors, the super script $\mathbf{h}^T$ denotes the transpose of vector $\mathbf{h}$, and $\mathbf{w}$ represents the model parameter.

Combined with Eq. (\ref{eq:sample}), the loss function of transductive learning is defined as:
\[
- \frac{1}{L} \sum_{i = 1}^L \log p(y_i | \mathbf{x}_i, \mathbf{e}_i) - \lambda \mathbb{E}_{(i, c, \gamma)} \log \sigma (\gamma \mathbf{w}_c^T \mathbf{e}_i),
\]
where the first term is defined by Eq. (\ref{eq:trans}), and $\lambda$ is a constant weighting factor.
The first term is the loss function of class label prediction and the second term is the loss function of context prediction.
This formulation is transductive because the prediction of label $y$ depends on the embedding~$\mathbf{e}$, which can only be learned for instances observed in the graph $A$ during training time.


\subsection{Inductive Formulation}

While we consider transductive learning in the above formulation, in many cases, it is desirable to learn a classifier that can generalize to unobserved instances, especially for large-scale tasks. For example, machine reading systems \cite{carlson2010toward} very frequently encounter novel entities on the Web and it is not practical to train a semi-supervised learning system on the entire Web. However, since learning graph embeddings is transductive in nature, it is not straightforward to do it in an inductive setting. Perozzi et al. \yrcite{perozzi2014deepwalk} addressed this issue by retraining the embeddings incrementally, which is time consuming and does not scale (and not inductive essentially).

To make the method inductive, the prediction of label $y$ should only depend on the input feature vector $\mathbf{x}$. Therefore, we define the embedding~$\mathbf{e}$ as a parameterized function of feature $\mathbf{x}$, as shown in Figure \ref{fig:ind}.
Similar to the transductive formulation, we apply $k$ layers on the input feature vector $\mathbf{x}$ to obtain $\mathbf{h}^k(\mathbf{x})$. However, rather than using a ``free'' embedding, we apply $l_1$ layers on the input feature vector $\mathbf{x}$ and define it as the embedding $\mathbf{e} = \mathbf{h}^{l_1}(\mathbf{x})$. Then another $l_2$ layers are applied on the embedding $\mathbf{h}^{l_2}(\mathbf{e}) = \mathbf{h}^{l_2}(\mathbf{h}^{l_1}(\mathbf{x}))$, denoted as $\mathbf{h}^l(\mathbf{x})$ where $l = l_1 + l_2$. The embedding $\mathbf{e}$ in this formulation can be viewed as a hidden layer that is a parameterized function of the feature $\mathbf{x}$.

With the above formulation, the label $y$ only depends on the feature $\mathbf{x}$. More specifically,
\begin{equation}
p(y | \mathbf{x}) = \frac{\exp [\mathbf{h}^k(\mathbf{x})^T, \mathbf{h}^l(\mathbf{x})^T] \mathbf{w}_y}{\sum_{y'} \exp [\mathbf{h}^k(\mathbf{x})^T, \mathbf{h}^l(\mathbf{x})^T] \mathbf{w}_{y'}}
\label{eq:ind}
\end{equation}

Replacing $\mathbf{e}_i$ in Eq. (\ref{eq:sample}) with $\mathbf{h}^{l_1}(\mathbf{x}_i)$, the loss function of inductive learning is
\[
- \frac{1}{L} \sum_{i = 1}^L \log p(y_i | \mathbf{x}_i) - \lambda \mathbb{E}_{(i, c, \gamma)} \log \sigma (\gamma \mathbf{w}_c^T \mathbf{h}^{l_1}(\mathbf{x}_i))
\]
where the first term is defined by Eq. (\ref{eq:ind}).


\subsection{Training}

We adopt stochastic gradient descent (SGD) \cite{bottou2010large} to train our model in the mini-batch mode.
We first sample a batch of labeled instances and take a gradient step to optimize the loss function of class label prediction. We then sample a batch of context $(i, c, \gamma)$ and take another gradient step to optimize the loss function of context prediction. We repeat the above procedures for $T_1$ and $T_2$ iterations respectively to approximate the weighting factor $\lambda$.
%
Algorithm \ref{algo:train} illustrates the SGD-based training algorithm for the transductive formulation. Similarly, we can replace $p(y_i | \mathbf{x}_i, \mathbf{e}_i)$ with $p(y_i | \mathbf{x}_i)$ in $\mathcal{L}_s$ to obtain the training algorithm for the inductive formulation. Let $\theta$ denote all model parameters. We update both embeddings $\mathbf{e}$ and parameters $\theta$ in transductive learning, and update only parameters $\theta$ in inductive learning. Before the joint training procedure, we apply a number of training iterations that optimize the unsupervised loss $\mathcal{L}_u$ alone and use the learned embeddings $\mathbf{e}$ as initialization for joint training.

\begin{algorithm}[tb]
    \caption{Model Training (Transductive)}
    \label{algo:train}
\begin{algorithmic}
    \STATE {\bfseries Input:} $A$, $\mathbf{x}_{1: L + U}$, $y_{1: L}$, $\lambda$, batch iterations $T_1, T_2$ and sizes $N_1, N_2$

    \REPEAT
        \FOR{$t \gets 1$ {\bfseries to} $T_1$}
            \STATE Sample a batch of labeled instances $i$ of size $N_1$
            \STATE $\mathcal{L}_s = - \frac{1}{N_1} \sum_i p(y_i | \mathbf{x}_i, \mathbf{e}_i)$
            \STATE Take a gradient step for $\mathcal{L}_s$
        \ENDFOR
        \FOR{$t \gets 1$ {\bfseries to} $T_2$}
            \STATE Sample a batch of context from $p(i, c, \gamma)$ of size $N_2$
            \STATE $\mathcal{L}_u = - \frac{1}{N_2} \sum_{(i, c, \gamma)} \log \sigma (\gamma \mathbf{w}_c^T \mathbf{e}_i)$
            \STATE Take a gradient step for $\mathcal{L}_u$
        \ENDFOR
    \UNTIL {stopping}
\end{algorithmic}
\end{algorithm}

\section{Experiments}
\label{sec:exp}

\begin{table}[tb]
\caption{Dataset statistics.}
\label{tab:stat}
\vskip 0.15in
\begin{center}
\begin{small}
\begin{sc}
\begin{tabular}{crrr}
Dataset & \#classes & \#nodes & \#edges \\
\hline
\abovespace
Citeseer & 6 & 3,327 & 4,732 \\
Cora & 7 & 2,708 & 5,429 \\
Pubmed & 3 & 19,717 & 44,338 \\
DIEL & 4 & 4,373,008 & 4,464,261 \\
NELL & 210 & 65,755 & 266,144 \\
\end{tabular}
\end{sc}
\end{small}
\end{center}
\vskip -0.1in
\end{table}

\begin{table}[tb]
\caption{Accuracy on text classification. Upper rows are inductive methods and lower rows are transductive methods.}
\label{tab:text}
\vskip 0.15in
\begin{center}
\begin{small}
\begin{sc}
\begin{tabular}{clll}
Method & Citeseer & Cora & Pubmed \\
\hline
\abovespace
Feat & 0.572 & 0.574 & 0.698 \\
ManiReg & 0.601 & 0.595 & 0.707 \\
SemiEmb & 0.596 & 0.590 & 0.711 \\
Planetoid-I & \textbf{0.647} & 0.612 & \textbf{0.772} \\
\hline
\abovespace
TSVM & 0.640 & 0.575 & 0.622 \\
LP & 0.453 & 0.680 & 0.630 \\
GraphEmb & 0.432 & 0.672 & 0.653 \\
Planetoid-G & 0.493 & 0.691 & 0.664 \\
Planetoid-T & 0.629 & \textbf{0.757} & 0.757 \\
\end{tabular}
\end{sc}
\end{small}
\end{center}
\vskip -0.1in
\end{table}

\begin{table}[tb]
\caption{Recall@$k$ on DIEL distantly-supervised entity extraction. Upper rows are inductive methods and lower rows are transductive methods. Results marked with $*$ are taken from the original DIEL paper \cite{bingimproving} with the same data splits.}
\label{tab:distant}
\vskip 0.15in
\begin{center}
\begin{small}
\begin{sc}
\begin{tabular}{cl}
Method & Recall@$k$ \\
\hline
\abovespace
$^*$Feat & 0.349 \\
ManiReg & 0.477 \\
SemiEmb & 0.486 \\
Planetoid-I & \textbf{0.501} \\
\hline
\abovespace
$^*$DIEL & 0.405 \\
$^*$LP & 0.162 \\
GraphEmb & 0.258 \\
Planetoid-G & 0.394 \\
Planetoid-T & \textbf{0.500} \\
\hline
\abovespace
$^*$Upper Bound & 0.617 \\
\end{tabular}
\end{sc}
\end{small}
\end{center}
\vskip -0.1in
\end{table}

\begin{table}[tb]
\caption{Accuracy on NELL entity classification with labeling rates of 
$0.1$, $0.01$, and $0.001$. Upper rows are inductive methods and lower rows are transductive methods.}
\label{tab:entity}
\vskip 0.15in
\begin{center}
\begin{small}
\begin{sc}
\begin{tabular}{clll}
Method & 0.1 & 0.01 & 0.001 \\
\hline
\abovespace
Feat & 0.621 & 0.404 & 0.217 \\
ManiReg & 0.634 & 0.413 & 0.218 \\
SemiEmb & 0.654 & 0.438 & 0.267 \\
Planetoid-I & 0.702 & 0.598 & 0.454 \\
\hline
\abovespace
LP & 0.714 & 0.448 & 0.265 \\
GraphEmb & 0.795 & 0.725 & 0.581 \\
Planetoid-G/T & \textbf{0.845} & \textbf{0.757} & \textbf{0.619} \\
\end{tabular}
\end{sc}
\end{small}
\end{center}
\vskip -0.1in
\end{table}

\begin{figure*}[tb]
\vskip 0.2in
\centering
\subfigure[GraphEmb]{\label{fig:deepwalk} \includegraphics[width = 0.32\textwidth]{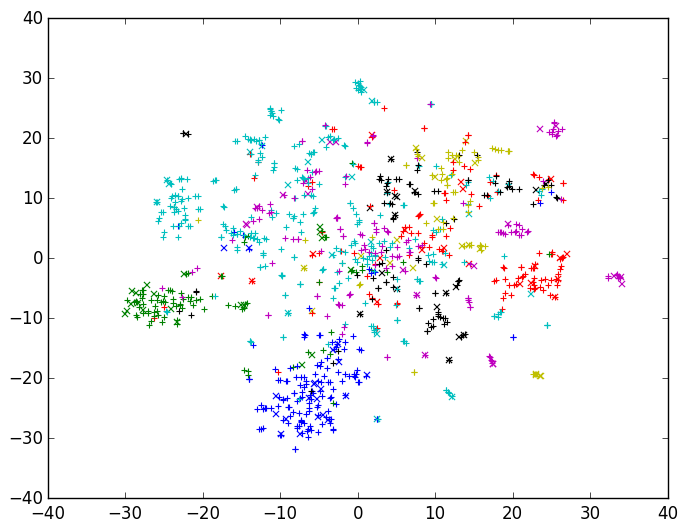}}
\subfigure[Planetoid-T]{\label{fig:pt} \includegraphics[width = 0.32\textwidth]{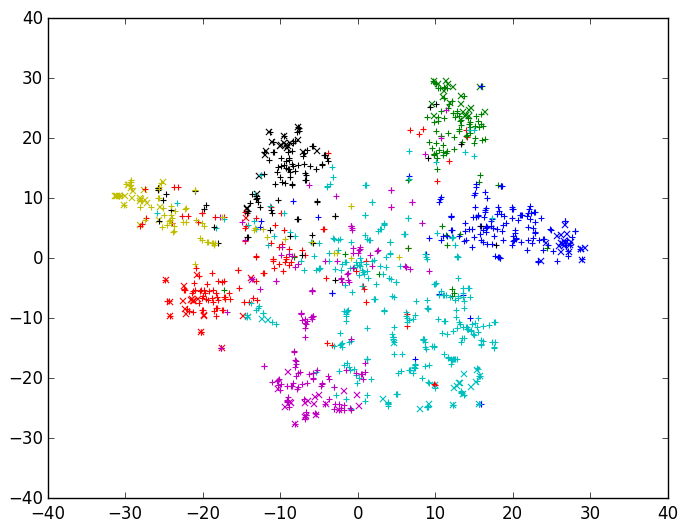}}
\subfigure[SemiEmb]{\label{fig:semi} \includegraphics[width = 0.32\textwidth]{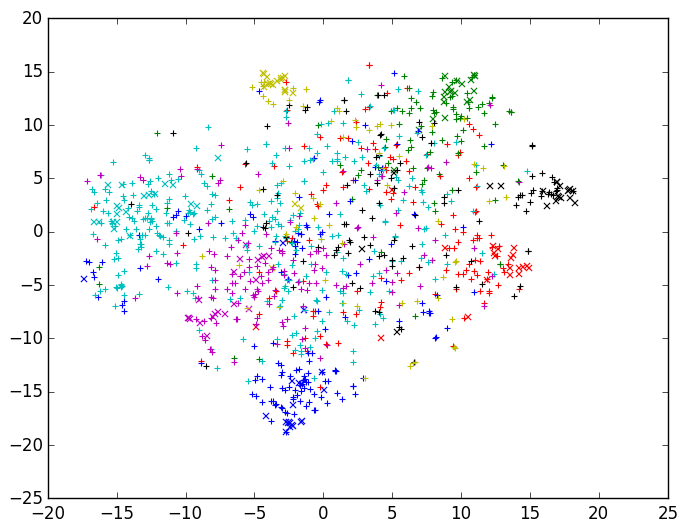}}
\caption{t-SNE Visualization of embedding spaces on the Cora dataset. Each color denotes a class.}
\label{fig:vis}
\vskip -0.2in
\end{figure*}

In our experiments, \textit{Planetoid-T} and \textit{Planetoid-I} denote the transductive and inductive formulation of our approach.
We compare our approach with label propagation (LP) \cite{zhu2003semi}, semi-supervised embedding (SemiEmb) \cite{weston2012deep}, manifold regularization (ManiReg) \cite{belkin2006manifold}, TSVM \cite{joachims1999transductive}, and graph embeddings (GraphEmb) \cite{perozzi2014deepwalk}.
Another baseline method, denoted as \textit{Feat}, is a linear softmax model that takes only the feature vectors $\mathbf{x}$ as input. We also derive a variant \textit{Planetoid-G} that learns embeddings to jointly predict class labels and graph context without use of feature vectors. The architecture of Planetoid-G is similar to Figure \ref{fig:trans} except that the \textit{input feature} and the corresponding hidden layers are removed.
Among the above methods, LP, GraphEmb and Planetoid-G do not use the features $\mathbf{x}$, while TSVM and Feat do not use the graph $A$. We include these methods into our experimental settings to better evaluate our approach.
Our preliminary experiments on the text classification datasets show that the performance of our model is not very sensitive to specific choices of the network architecture\footnote{We note that 
it is possible to develop other architectures for different 
applications, such as using a shared hidden layer for feature vectors and embeddings.}.
We adapt the implementation of GraphEmb\footnote{\url{https://github.com/phanein/deepwalk}} to our Skipgram implementation. We use the Junto library \cite{talukdar2009new} for label propagation, and SVMLight\footnote{\url{http://svmlight.joachims.org/}} for TSVM. We also use our own implementation of ManiReg and SemiEmb by modifying the symbolic objective function in Planetoid. In all of our experiments, we set the model hyper-parameters to $r_1 = 5/6$, $q = 10$, $d = 3$, $N_1 = 200$ and $N_2 = 200$ for Planetoid.
We use the same $r_1$, $q$ and $d$ for GraphEmb, and the same $N_1$ and $N_2$ for ManiReg and SemiEmb. We tune $r_2$, $T_1$, $T_2$, the learning rate and hyper-parameters in other models based on an additional data split with a different random seed.

The statistics for five of our benchmark datasets are reported in Table \ref{tab:stat}.
For each dataset, we split all instances into three parts, labeled data, unlabeled data, and test data. Inductive methods are trained on the labeled and unlabeled data, and tested on the test data. Transductive methods, on the other hand, are trained on the labeled, unlabeled data, and test data without labels.


\subsection{Text Classification}

We first considered three text classification datasets\footnote{\url{http://linqs.umiacs.umd.edu/projects//projects/lbc/}}, Citeseer, Cora and Pubmed \cite{sen2008collective}. Each dataset contains bag-of-words representation of documents and citation links between the documents. We treat the bag-of-words as feature vectors $\mathbf{x}$. We construct the graph $A$ based on the citation links; if document $i$ cites $j$, then we set $a_{ij} = a_{ji} = 1$. The goal is to classify each document into one class.
We randomly sample $20$ instances for each class as labeled data, $1,000$ instances as test data, and the rest are used as unlabeled data. The same data splits are used for different methods, and we compute the average accuracy for comparison.

The experimental results are reported in Table \ref{tab:text}. Among the inductive methods, Planetoid-I achieves the best performance on all the three datasets with the improvement of up to $6.1\%$ on Pubmed, which indicates that our embedding techniques are more effective than graph Laplacian regularization.
Among the transductive methods, Planetoid-T achieves the best performance on Cora and Pubmed, while TSVM performs the best on Citeseer. However, TSVM does not perform well on Cora and Pubmed. 
Planetoid-I slightly outperforms Planetoid-T on Citeseer and Pubmed, while Planetoid-T gets up to $14.5\%$ improvement over Planetoid-I on Cora. We conjecture that in Planetoid-I, the feature vectors impose constraints on the learned embeddings, since they are represented by a parameterized function of the input feature vectors. If such constraints are appropriate, as is the case on Citeseer and Pubmed, it improves the non-convex optimization of embedding learning and leads to better performance. However, if such constraints rule out the optimal embeddings, the inductive model will suffer.

Planetoid-G consistently outperforms GraphEmb on all three datasets, which indicates that joint training with label information can improve the performance over training the supervised and unsupervised objectives separately.
Figure~\ref{fig:vis} displays the $2$-D embedding spaces on the Cora dataset
using t-SNE~\cite{van2008visualizing}. Note that  
different classes are better separated in the embedding space of Planetoid-T than that of GraphEmb and SemiEmb, which is consistent with our empirical findings. We also observe similar results for the other two datasets.  


\subsection{Distantly-Supervised Entity Extraction}

We next considered the DIEL (Distant Information Extraction using coordinate-term Lists) dataset \cite{bingimproving}.
The DIEL dataset contains pre-extracted features for each entity mention in text, and a graph that connects entity mentions to coordinate lists. The goal is to extract medical entities from text given feature vectors and the graph.

We follow the exact experimental setup as in the original DIEL paper \cite{bingimproving}, including data splits of different runs, preprocessing of entity mentions and coordinate lists, and evaluation. We treat the top-$k$ entities given by a model as positive instances, and compute recall@$k$ for evaluation ($k$ is set to $240,000$ following the DIEL paper). We report the average result of 10 runs in Table \ref{tab:distant}, where \textit{Feat} refers to a result obtained by SVM (referred to as DS-Baseline in the DIEL paper). The result of LP was also taken from~\cite{bingimproving}. \textit{DIEL} in Table \ref{tab:distant} refers to the method proposed by the original paper, which is an improved version of label propagation that trains classifiers on feature vectors based on the output of label propagation. We did not include TSVM into the comparison since it does not scale.
Since we use Freebase as ground truth and some entities are not present in text, the upper bound of recall as shown in Table \ref{tab:distant} is $0.617$.

Both Planetoid-I and Planetoid-T significantly outperform all other methods.
Each of Planetoid-I and Planetoid-T achieves the best performance in 5 out of 10 runs, and they give a similar recall on average, which indicates that there is no significant difference between these two methods on this dataset. Planetoid-G clearly outperforms GraphEmb, which again shows the benefit of joint training.

\subsection{Entity Classification}

We sorted out an entity classification dataset from the knowledge base of Never Ending Language Learning (NELL)
\cite{carlson2010toward} and a hierarchical entity classification dataset \cite{dalvi2014hierarchical}
that links NELL entities to text in ClueWeb09. We extracted the entities and the relations between entities from the NELL knowledge base, and then obtained text description by linking the entities to ClueWeb09. We use text bag-of-words representation as feature vectors of the entities.

We next describe how to construct the graph based on the knowledge base. We first remove relations that are not populated in NELL, including ``generalizations'', ``haswikipediaurl'', and ``atdate''. In the knowledge base, each relation is denoted as a triplet $(e_1, r, e_2)$, where $e_1$, $r$, $e_2$ denote head entity, relation, and tail entity respectively. We treat each entity $e$ as a node in the graph, and each relation $r$ is split as two nodes $r_1$ and $r_2$ in the graph. For each $(e_1, r, e_2)$, we add two edges in the graph, $(e_1, r_1)$ and $(e_2, r_2)$.

We removed all classes with less than $10$ entities. The goal is to classify the entities in the knowledge base into one of the $210$ classes given the feature vectors and the graph. Let $\beta$ be the labeling rate. We set $\beta$ to $0.1$, $0.01$, and $0.001$. $\max(\beta N, 1)$ instances are labeled for a class with $N$ entities, so each class has at least one entity in the labeled data.

We report the results in Table \ref{tab:entity}. We did not include TSVM since it does not scale to such a large number of classes with the one-vs-rest scheme. Adding feature vectors does not improve the performance of Planetoid-T, so we set the feature vectors for Planetoid-T to be all empty, and therefore Planetoid-T is equivalent to Planetoid-G in this case.

Planetoid-I significantly outperforms the best of the other compared inductive methods---i.e., SemiEmb---by $4.8\%$, $16.0\%$, and $18.7\%$ respectively with three labeling rates. As the labeling rate decreases, the improvement of Planetoid-I over SemiEmb becomes more significant.

Graph structure is more informative than features in this dataset, so inductive methods perform worse than transductive methods. Planetoid-G outperforms GraphEmb by $5.0\%$, $3.2\%$ and $3.8\%$.


\section{Conclusion}
\label{sec:conc}

Our contribution is three-fold: a) incontrast to previous semi-supervised learning approaches that largely depend on graph Laplacian regularization, we propose a novel approach by joint training of classification and graph context prediction; b) since it is difficult to generalize graph embeddings to novel instances, we design a novel inductive approach that conditions embeddings on input features; c) we empirically show substantial improvement over existing methods (up to $8.5\%$ and on average $4.1\%$), and even more significant improvement in the inductive setting (up to $18.7\%$ and on average $7.8\%$).


Our experimental results on five benchmark datasets also show that a) joint training gives improvement over unsupervised learning; b) predicting graph context is more effective than graph Laplacian regularization; c) the performance of the inductive variant depends on the informativeness of feature vectors. 

One direction of future work would be to 
apply our framework to more complex networks, including recurrent networks. It would also be interesting to experiment with datasets where a graph is computed based on distances between feature vectors.

\section*{Acknowledgements}

This work was funded by the NSF under grants CCF-1414030 and IIS-1250956, and by Google.

\bibliography{main}
\bibliographystyle{icml2016}

\end{document}